\documentclass{article}
\usepackage{ijcai23}

\usepackage{times}
\usepackage{soul}
\usepackage{url}
\usepackage[hidelinks]{hyperref}
\usepackage[utf8]{inputenc}
\usepackage[small]{caption}
\usepackage{graphicx}
\usepackage{amsmath}
\usepackage{amsthm}
\usepackage{booktabs}
\usepackage[linesnumbered,ruled]{algorithm2e}

\usepackage[switch]{lineno}
\usepackage{xcolor}
\usepackage{hyperref}

\newtheorem{example}{Example}
\newtheorem{theorem}{Theorem}

\title{Tracking Players in a Badminton Court by Two Cameras
}

\author{
Young-Ching Chou$^1$\and
Shen-Ru Zhang$^1$\and
Bo-Wei Chen$^1$\and
Hong-Qi Chen$^1$\and
Cheng-Kuan Lin$^{1,2,3}$\And
Yu-Chee Tseng$^1$
\affiliations
$^1$ Department of Computer Science, National Yang Ming Chiao Tung University, Taiwan\\
$^2$ Undergraduate Degree Program of Systems Engineering and Technology, National Yang Ming Chiao Tung University, Taiwan\\
$^3$ Computer Science and Information Engineering, Chung Cheng Institute of Technology, NDU, Taiwan\\
\emails
melody.c@nycu.edu.tw, a1301145@gmail.com, brian1357929@gmail.com, l311551105.cs11@nycu.edu.tw, \{cklin,yctseng\}@cs.nycu.edu.tw
}

\begin{document}

\maketitle

\begin{abstract}
This study proposes a simple method for multi-object tracking (MOT) of players in a badminton court. We leverage two off-the-shelf cameras, one on the top of the court and the other on the side of the court. The one on the top is to track players' trajectories, while the one on the side is to analyze the pixel features of players. By computing the correlations between adjacent frames and engaging the information of the two cameras, MOT of badminton players is obtained. This two-camera approach addresses the challenge of player occlusion and overlapping in a badminton court, providing player trajectory tracking and multi-angle analysis. The presented system offers insights into the positions and movements of badminton players, thus serving as a coaching or self-training tool for badminton players to improve their gaming strategies.
\end{abstract}


\section{Introduction}
Position and trajectory tracking of players and opponents is crucial in any ball game \cite{hu2004}, \cite{la2007}, \cite{od2009}, \cite{hsu2019coachai}, \cite{tai2019toward}. It can be further extended to movement patterns analysis. Badminton is one of the popular sports worldwide. Through interviews with badminton experts, coaches, and commentators, it becomes evident that players positioning significantly impacts gaming performance. Therefore, in badminton matches, observing the positioning of opponent(s) provides important clues for tactics and strategies. Such an analysis helps understand the strengths, weaknesses, and potential reactions of players, especially in evaluating whether to attack or defend. For example, if an opponent is near the baseline, one can choose to play a short shot or increase the badminton's height to make it difficult for the opponent to return. On the other hand, observing an opponent's movement patterns can reveal his/her preferences and habits, based on which a player can select the direction of a serve or the angle of an attack. Therefore, it is desirable to have an efficient system to conduct badminton players' positions and moving trajectories during a match.

In a video recording of sports, multiple object tracking (MOT) is an important task \cite{cui2023}. MOT involves object detection and track maintenance \cite{wa2021}. In practice, occlusion and overlapping of objects have always been challenging issues \cite{zh2016}. Even when tracking individuals, accurately identifying and associating targets become difficult when they are occluded or overlapped \cite{wa2022}. These challenges become even more difficult to handle during a badminton match because players usually make very fast back-and-forth movements.

\begin{figure}[htbp]
   \centerline{\includegraphics[width=3.0in]{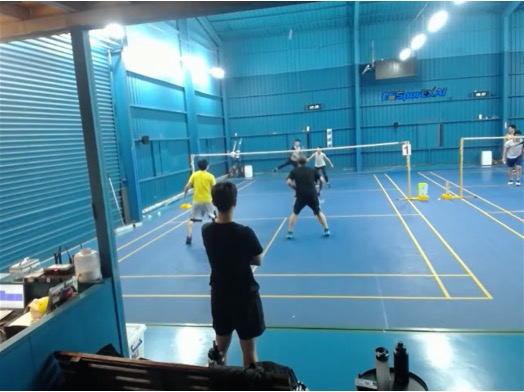}}
  \caption{The perspective of the badminton court image that does not conform to the requirements of match recording.}
  \label{fig:fig1}
\end{figure}

\begin{figure}[t]
   \centerline{\includegraphics[width=3.0in]{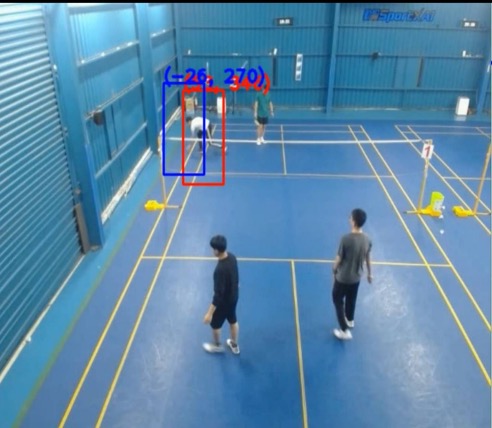}}
  \caption{Incorrect recognition caused by overlapping.}
  \label{fig:fig2}
\end{figure}

With rapid technological advancements, there has been significant progress in the quality of match broadcasts, allowing viewers to enjoy the game with clarity. However, inappropriate camera angles can cause significant obstruction of the view and analysis of the match by equipment and personnel at the venue, as shown in Figure \ref{fig:fig1}. Furthermore, due to limitations in camera angles on the court, it remains challenging to avoid the problem of certain exciting moments being obstructed by players or equipment. This leads to insufficient effectiveness in intelligent match analysis when relying on a single camera, as depicted in Figure \ref{fig:fig2}. This study proposes a two-camera system for analyzing badminton player positioning and movement patterns, providing an effective and low-cost solution for tracking player trajectories and presenting strategic evaluation information for positioning analysis. This system facilitates the selection of appropriate tactics and strategies for badminton players.

\section{Related Work}

 In the various video footage obtained from different badminton venues, our goal is to provide analysis that enables badminton players to understand their own positions, the positions of their opponents, and the movement patterns. Therefore, we explore methods in object tracking research for connecting the detection results of each frame to form trajectories. The following are commonly used methods and relevant literature:

 \begin{enumerate}
    \item {\bf Kalman Filter:} The Kalman filter is a commonly used method for estimating system states and is widely applied in track maintenance for object tracking. It predicts and updates the target trajectory by considering the current observations and previous states. In the study by Chen et al. \cite{ch2019}, a hardware solution utilizing lidar and a software solution utilizing support vector machines (SVM) were used to build a leg classifier for human tracking, employing the Kalman filter algorithm. However, considering commercial factors, we need to use cost-effective cameras instead of the higher-cost ones used in the referenced study. Nevertheless, the concept of SVM classifiers proposed in the literature serves as a valuable consideration for our software recognition approach.
    \item  {\bf Association-based Methods:} This method utilizes information such as appearance, motion, and spatial positions of targets to establish associations between them and connect trajectories. Berclaz et al. propose a multi-camera pedestrian tracking method which can generate correct trajectories even when a single target is not detected in a frame by considering the previous and subsequent frames \cite{be2011}. This research inspired our idea of installing overhead cameras to limit the range of associations for recognizing the movement trajectories of badminton players.
    \item {\bf Data Association:} This method involves matching and associating targets in each frame based on factors such as appearance similarity, motion consistency, and distance metrics \cite{mi2016}. This idea prompts us to consider the use of Histogram of Oriented Gradients (HOG) features and distance metrics for detecting target associations.
     \item  {\bf Graph-based Methods:} These methods model the target tracking problem as an optimization problem on graphs and utilize graph theory algorithms to connect trajectories. In the study by Xie and Sun \cite{xi2023}, a graph-based top-down visual attention model was constructed using multi-scale transformations. We conceived the idea of visualizing player movement through a visual model.
     \item  {\bf Deep Learning Methods:} These methods extract features and use convolutional neural networks (CNN) or recurrent neural networks (RNN) to learn target representations and connect trajectories. Current research in this area primarily focuses on addressing occlusion and overlapping issues \cite{zh2023}.
\end{enumerate}

Considering that badminton courts involve multiple object tracking due to doubles matches, the issues of occlusion and overlapping (as shown in Figure \ref{fig:fig2}) need to be addressed. In terms of hardware, we conducted experiments by installing overhead cameras at a bird's-eye angle, and for maintaining player trajectories, we employed feature-based object detection within a certain distance range and connected the detected targets from different frames to visualize the trajectories.

\begin{figure}[t]
   \centerline{\includegraphics[width=3.3in]{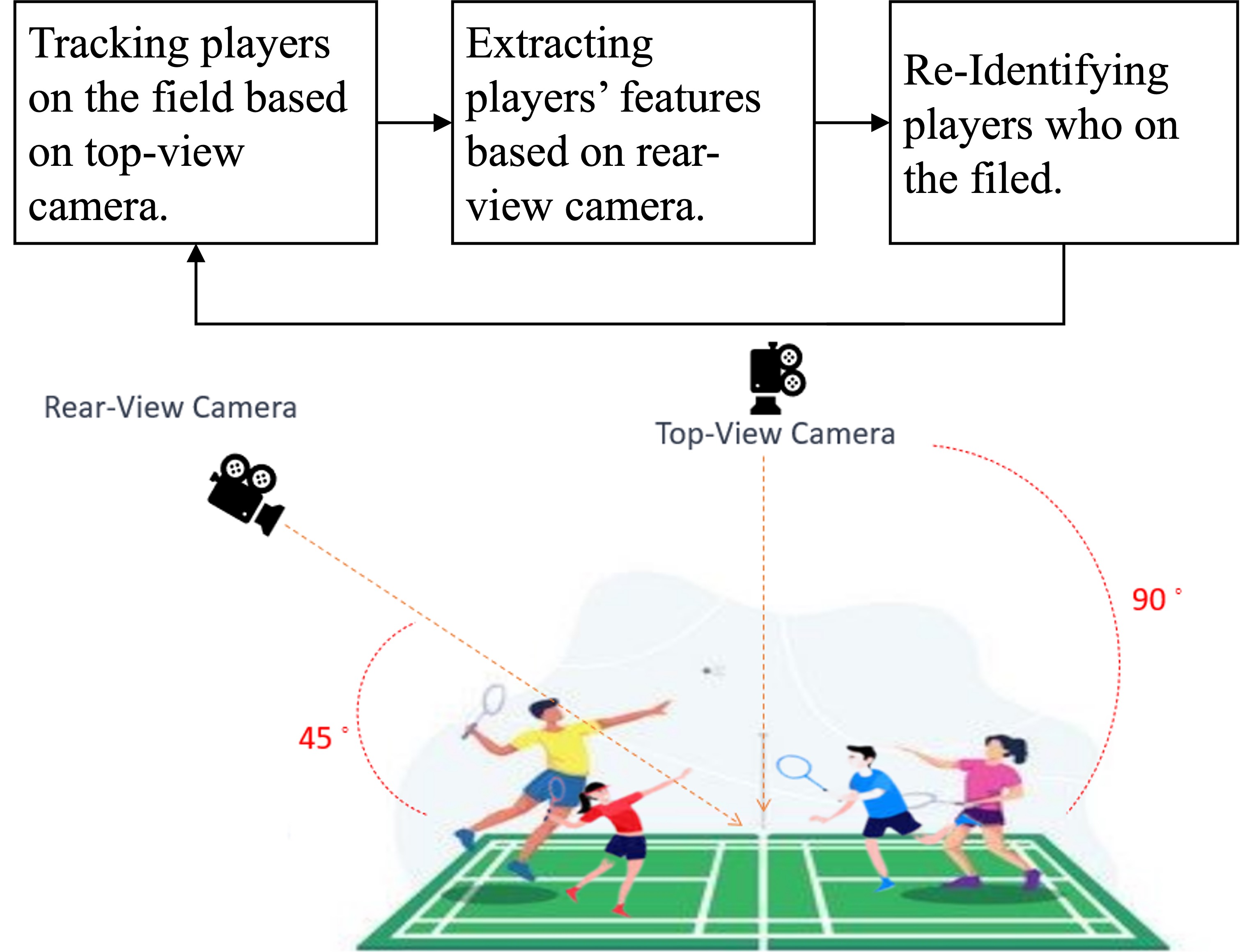}}
  \caption{Camera coordination architecture diagram.}
  \label{fig:mainStruc}
\end{figure}

\section{Proposed Two-Camera Tracking System}
Badminton is a fast-paced sport, and therefore the issue of player overlap persists during competitions. An example is shown in Figure \ref{fig:fig2}. To address this problem, we propose a two-camera configuration as depicted in Figure \ref{fig:mainStruc}. The top-view camera is utilized to provide an overhead perspective, aiming to tackle obstructions and overlapping. Conversely, the side-view camera, resembling conventional sports broadcast angles, captures players' body features. The camera coordination and ReID (Re-Identification) module functions aim to effectively mitigate these issues by integrating information from both camera angles.
Figure \ref{fig:mainStruc} illustrates an example of how this solution resolves the overlapping issue depicted in Figure \ref{fig:fig2}. 
We utilize the top-view perspective to handle players positioning and trajectory tracking. 
While in most cases, the top-view camera can establish players trajectories, the information obtained from the top-view camera is unable to handle the reconstruction of interrupted trajectories caused by players entering or exiting the field.
When players exit or enter the field, we switch to the rear-view perspective to identify the players and process ReID to track all players trajectories.

\begin{figure}[htbp]
   \centerline{\includegraphics[width=3.5in]{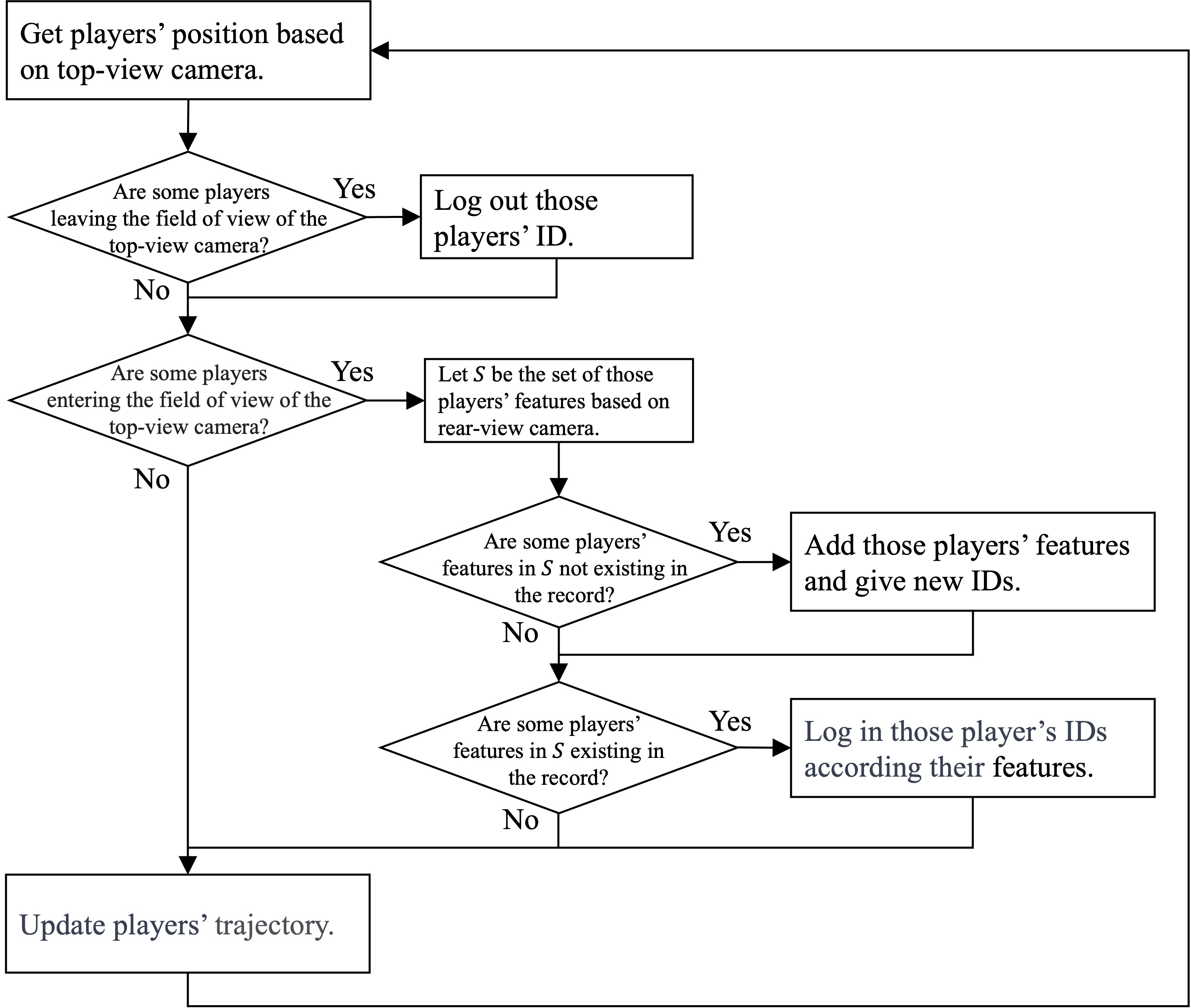}}
  \caption{Flowchart of algorithm \ref{alg:algorithm}}
  \label{fig:flowalgo}
\end{figure}

Algorithm \ref{alg:algorithm} shows the detail of our method how to identify players position and how to ReID. It aims to identify the players' ID, ID state, and features using two-view cameras. The input consists of a list of player information, $listPlayersInfo$ containing the player's ID ($listPlayersInfo.id$), ID login state ($listPlayersInfo.idState$), feature ($listPlayersInfo.feature$), and positions ($listPlayersInfo.position$). The output is an updated list of player information, $listPlayersInfo$.
The algorithm starts by obtaining the players' positions based on the top-view camera. If any players are detected to have left the field of view of the top-view camera, their respective $listPlayersInfo.idState[i]$ is set to "logout".
Next, the algorithm checks if any players have entered the field of view of the top-view camera. If there are new players detected, their features are obtained from the rear-view camera (stored in $S$). The algorithm iterates through the elements in $S$.
For each element $s$ in $S$, the algorithm compares it with the existing features in $listPlayersInfo$. If s does not match any existing $listPlayersInfo.feature[i]$ for all $i$, it indicates a new player. In this case, the player's features are added to $listPlayersInfo$, and a new ID is assigned.
On the other hand, if s matches an existing $listPlayersInfo.feature[i]$ for some $i$, it implies that the player is already present in $listPlayersInfo$. In this case, the $listPlayersInfo.idState[i]$ is set to ``login''.
Finally, the algorithm returns the updated $listPlayersInfo$. Figure \ref{fig:flowalgo} shows the flowchart of Algorithm \ref{alg:algorithm}. 

In order to address these challenges, we will now provide an explanation of our ideas and technical approaches for each functional module.

\begin{algorithm}
   \caption{Identify Players' ID, state of ID, and Features based on Two-View Cameras}
   \label{alg:algorithm}
    \SetKwInOut{Input}{Input}
    \SetKwInOut{Output}{Output}

    \Input{$listPlayersInfo$ which contains Player's ID, ID login state, feature, positions}
    \Output{$listPlayersInfo$}
    Get players’ position based on top-view camera;\\
    \If{there are some players leaving the field of view of the top-view camera}{
        Set those players' $listPlayersInfo.idState[i]$ as logout;
    }
    \If{there are some players entering the field of view of the top-view camera}{
       $S \leftarrow$ those players’ features based on rear-view camera\;
       While{$S == \emptyset$}{
            Choose an element $s$ form $S$;\\
            \If{$s \neq listPlayersInfo.feature[i]$ for each $i$}{
                 Add this player's features and give new IDs to $listPlayersInfo$;\\
            }
            \ElseIf{$s == listPlayersInfo.feature[i]$ for some $i$}{
            Set $listPlayersInfo.idState[i]$ as login;\\
            }  
       }
    }
    \Return{$listPlayersInfo$};
\end{algorithm}

\subsection{Phase 1: Track players on the field}

\begin{figure}[t]
  \centerline{\includegraphics[width=3.3in]{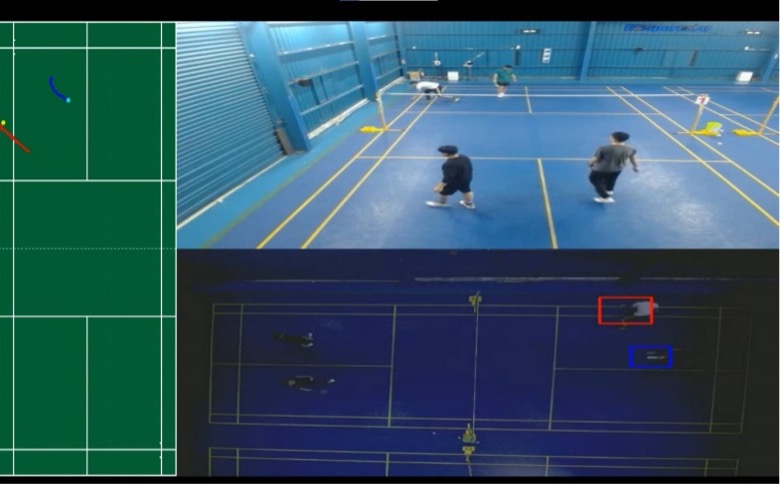}}
  \caption{The effective resolution of overlapping and occlusion issues through the methodology proposed in this study.}
  \label{fig:fig4}
\end{figure}

In this phase, we utilize a top-view camera to acquire the positions of players on the field and track the players' trajectory.
The top-view module involves positioning the camera above the badminton court with a downward angle of $90$ degrees towards the ground. The purpose of the top-view camera is to track player trajectories and prevent obstruction and overlapping within the field of view. 
To improve the accuracy and robustness of target tracking, we utilize the Channel and Spatial Reliability Tracking (CSRT) technique \cite{lu2017}. This algorithm combines color channel and spatial information to enhance target tracking performance. 
In Figure \ref{fig:fig4}, CSRT creates player trackers (tracker1 represented by a red box and tracker2 represented by a blue box) on the half badminton court captured by the top-view camera.

\subsection{Phase 2: Extract player features}
We use rear-view module is to identify the  features of players. This camera is mounted above and to the rear, capturing images at a downward angle of $45$ degrees. This angle provides the optimal perspective for capturing clear visual features of individuals. 

We employ the Histogram of Oriented Gradients (HOG) technique to describe image features for image processing and computation. In the rear-view module, we use the video footage captured from the rear-view camera as input and apply the HOG technique to extract the appearance information of players. The goal is to detect the body features specific to  players. We incorporate a Support Vector Machine (SVM) detector \cite{sh2022} into the system. Specifically, SVM is a supervised learning algorithm that classifies whether a region in an image is the target region or a non-target region in object detection tasks. In our case, HOG is used to extract features (images or features within images), which are then input to the SVM detector to classify whether they represent the features of badminton players. SVM optimizes the position and shape of the hyperplane by minimizing the objective function, aiming to separate the target and non-target regions as effectively as possible. The output of the rear-view module is the position coordinates of the identified badminton players on the court captured by the rear-view camera.

\subsection{Phase 3. ReID}
When the trajectory of a player approaches or crosses the predefined boundary of the court for an extended period, resulting in a sudden interruption or disappearance of the trajectory, we activate the ReID model by coordinating the two camera angles.
The ReID model is proposed to explain its operation:


\begin{enumerate}
   \item {\bf Identification of individuals:} The ReID model aims to identify individuals who have left the court based on their appearance. It takes as input the images or features extracted from the images captured by the rear-view and top-view cameras. By comparing these features with a database of known individuals, the ReID model determines whether the identified individuals match any of the players on the record or if they are new individuals.
   \item  {\bf Matching algorithm:} To perform the identification process, the ReID model utilizes a matching algorithm that compares the features of the unidentified individuals with the features of the players on the record. It measures the similarity or dissimilarity between the features and assigns a matching score.
   \item {\bf Decision-making:} Based on the matching scores, the ReID model makes a decision regarding the identity of the unidentified individuals. If the matching score exceeds a certain threshold and matches one of the  players on the record, the ReID model assigns the corresponding original ID to the identified individual. If the matching score is below the threshold or does not match any of the original four players, the ReID model assigns a new ID to the unidentified individual.
   \item {\bf Updating player information:} After identifying the individuals who have left the court, the ReID model updates their corresponding IDs, features, and positions. This ensures that their trajectories can be correctly tracked and recorded in subsequent frames.
\end{enumerate}

By incorporating the ReID model, we can handle scenarios where players' trajectories are interrupted or disappear due to crossing the predefined court boundaries. The model enables the system to identify and track individuals accurately, maintaining consistency in player identification throughout the game.

The difficulty of ReID lies in how to combine the information of top-view and rear-view to confirm the identity of players, especially when players repeatedly enter and exit the  filed.

Based on the rear-view camera and top-view camera, we need to perform perspective transformation on the obtained player positions from different cameras. Our perspective transformation involves converting the images from different camera angles to world coordinates based on the known real dimensions of the badminton court. The rear-view module is used to transform the recognized player positions in rear-view camera coordinates to world coordinates, and then pass the position information to the top-view module. The top-view module transforms the world coordinates into top-view camera coordinates to track the positions and trajectories of the players. The track output model takes the world coordinates and presents the positions of the players from both camera angles on the screen. In this study, we apply perspective transformation techniques to convert coordinates from one camera angle to another, and visualize the transformed data on a standardized badminton court to display the trajectory results. To achieve player position annotation and trajectory implementation in different camera angles, we utilize the perspective transformation matrices and coordinate transformation operations.
In the following, we provide an explanation of the mathematical models for the position coordinates on the badminton court in the two camera angles:

\begin{enumerate}
    \item{\bf Perspective transformation model for coordinate conversion:}
    The purpose of coordinate conversion is to accurately transform and measure coordinates between different camera angles using perspective transformation. We have two sets of coordinates for different camera angles:
    \begin{enumerate}
        \item {\bf Rear-view module with $Side_{pts1}$ and $Side_{pts2}$ coordinate matrices:} $Side_{pts1}$ represents the coordinates of four selected points in the rear-view video, corresponding to the four corners of the badminton court in the rear-view video. $Side_{pts2}$ represents the coordinates of the four points in the world coordinate system, corresponding to the known standard corners of the badminton court. The purpose of $Side_{pts1}$ and $Side_{pts2}$ is to transform the badminton court in the rear-view video to a standard rectangle or quadrilateral in the world coordinate system through perspective transformation, enabling coordinate alignment and measurement.
        \item {\bf Top-view module with $Roof_{pts1}$ and $Roof_{pts2}$ coordinate matrices:} $Roof_{pts1}$ represents the coordinates of four selected points in the world coordinate system. $Roof_{pts2}$ represents the coordinates of the four points in the top-view video, corresponding to the corners of the badminton court. The purpose of $Roof_{pts1}$ and $Roof_{pts2}$ is to transform the rectangle or quadrilateral in the top-view video to a standard rectangle or quadrilateral in the world coordinate system through perspective transformation, enabling coordinate alignment and measurement.
    \end{enumerate}
    In summary, $Side_{pts1}$, $Side_{pts2}$, $Roof_{pts1}$, and $Roof_{pts2}$ are selected point locations used for perspective transformation between different camera angles, enabling accurate coordinate conversion and measurement between different coordinate systems.

    \item{\bf Computation of perspective transformation matrices:}
    We use the $cv2.getPerspectiveTransform$ function to obtain the perspective transformation matrices. By using the point coordinates from the rear-view module's rear-view video and the corresponding point coordinates in the world coordinate system as input, we calculate the perspective transformation matrix $Side_M$. Similarly, for the top-view module, we calculate the perspective transformation matrix $Roof_M$ for $Roof_{pts1}$ and $Roof_{pts2}$. The Track Output Model uses the inverse matrix of $Roof_M$, $Roof_{M2}$, to transform the top-view camera coordinate system back to the world coordinate system for displaying tracking trajectories on a standardized plane diagram.
    \item{\bf Coordinate conversion formula for rear-view module:}
    $Side_{pts1}$ represents the coordinates of four selected points in the rear-view video. $Side_{pts2}$ represents the coordinates of the four points in the world coordinate system corresponding to $Side_{pts1}$. $Side_M$ is the perspective transformation matrix used to convert the points from the rear-view video's image coordinate system to the world coordinate system. It can be expressed as:
        $$[x', y', 1'] = Side_M[x, y, l]$$
    where $[x, y, 1]$ represents the point coordinates in the rear-view perspective video, and $[x', y', w']$ represents the point coordinates in the world coordinate system.
     
    \item{\bf Coordinate conversion formula for top-view module:}
    $Roof_{pts1}$ represents the coordinates of four selected points in the top-view perspective video. $Roof_{pts2}$ represents the coordinates of the corresponding four points in the world coordinate system. $Roof_M$ is the perspective transformation matrix that converts points from the image coordinate system in the top-view perspective video to the world coordinate system. It can be represented as:
       $$[x', y', w'] = Roof_M[x, y, 1]$$
    where $[x, y, 1]$ represents the point coordinates in the top-view perspective video, and $[x', y', w']$ represents the point coordinates in the world coordinate system.

    \item{\bf Coordinate conversion formula for track output model:}  
    $Roof_{M2}$ is the inverse matrix of $Roof_M$. Its purpose is to convert the trajectory points from the top-view video image coordinate system to the world coordinate system. It can be expressed as: 
    $$[x', y', w'] = Side_{M_2}[x, y, l]$$
    where $[x, y, 1]$ represents the coordinates of the moving trajectory points in the top-view perspective video, and $[x', y', w']$ represents the coordinates in the world coordinate system. Through the aforementioned mathematical model and perspective transformation matrices, the coordinates of points in the rear-view and top-view perspective videos can be transformed into the world coordinate system, enabling coordinate alignment and trajectory visualization.
    The purpose of these steps is to achieve coordinate alignment and measurement by transforming the coordinates of points from the rear-view and top-view perspective videos into the world coordinate system, facilitating further processing and visualization in a unified coordinate system.
\end{enumerate}

\begin{figure}[t]
  \centerline{\includegraphics[width=3.3in]{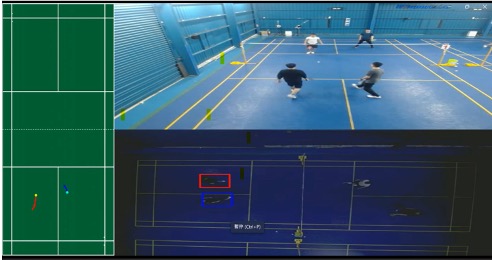}}
  \caption{The visual representation of two-camera perspectives, player positions, and movement trajectories achieved in this study.}
  \label{fig:fig3}
\end{figure}

\section{Results}
We employed human detectors and trackers, combined with perspective transformation and trajectory visualization, to achieve player tracking and analysis in a badminton match. (As shown in Figure \ref{fig:fig3})

Through the design of this study, combining the rear-view and top-view perspectives with perspective transformation for comparison, the problem of tracking badminton players on the court with occlusion and overlapping has been effectively addressed (see Figure \ref{fig:fig4}). Figure \ref{fig:fig2} and Figure \ref{fig:fig4} depict the same video segment. Prior to incorporating the top-view perspective through perspective transformation, it can be observed that due to the occurrence of overlapping zones, both badminton players were mistakenly identified as the player wearing a white shirt and bending over in Figure \ref{fig:fig2}. However, in Figure \ref{fig:fig4}, through the comparison with perspective transformation and reference, the movement trajectories of both players are clearly depicted. The wrongly recognized player in the white shirt and bending posture in Figure \ref{fig:fig2} is represented by the red movement trajectory on the flat diagram of the badminton court, while the initially unrecognized player is represented by the blue movement trajectory. Therefore, this study effectively and economically solves the problem of occlusion and overlapping.

\begin{figure}[t]
  \centerline{\includegraphics[width=3.3in]{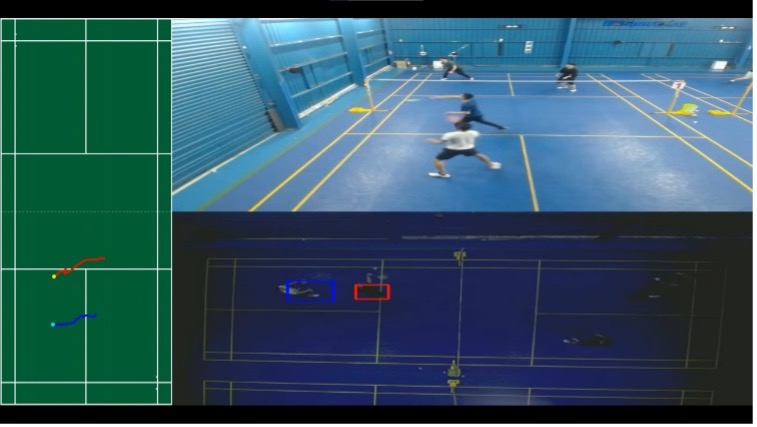}}
  \caption{The occurrence of scoring errors due to incorrect positioning and movement trajectories.}
  \label{fig:fig5}
\end{figure}

The primary objective of this study is to provide badminton players with the ability to observe their own and their opponents' positions on the court: by analyzing the opponent's position, they can choose between offense and defense. Figure \ref{fig:fig5} shows that both players on the court are positioned on the same side, indicating a lack of even defense, which results in losing points. Therefore, through the visual presentation provided by this system, players can truly understand the correctness of their positions and delve into the discussion on movement trajectory. This serves as a basis for improving tactics in terms of subsequent attacks or defense for badminton players and identifying areas for further training.

\section{Discussion}
In this study, dynamic trajectories were utilized to track individuals in the video. The individuals were marked with colored rectangular bounding boxes. Moreover, the system establishes connections between the players' movement positions and illustrates their trajectories on a standardized layout of the badminton court in consecutive frames, with each player's trajectory represented by a designated color.This visual representation allows badminton players to easily compare and analyze their own and their opponents' movements and positioning on the court. It also provides a more intuitive understanding of the badminton match proceedings for commentators and spectators. We provide a link to video clips processed by our system for reference: \url{https://drive.google.com/file/d/13iKaklXw3a07TWJZ9iy9r5NeHbAgx3fj/view?usp=share_link}

In future research, we aim to further enhance our understanding of the relationship between positioning, movement analysis, and the scoring outcome by incorporating intelligent scoring systems. Additionally, we strive to develop the concept of a smart badminton court. It is our hope that through our research, we can assist badminton players in improving their skills and abilities. Furthermore, we envision that the  of our system will evolve to become valuable references for badminton training modes and strategic decision-making.

\bibliographystyle{named}
\bibliography{ijcai23}
\end{document}